# *Digital me ontology and ethics*

Ljupco Kocarev [*] and Jasna Koteska [**]

21 December 2020

[*] Macedonian Academy of Sciences and Arts, Skopje, Macedonia[1], lkocarev@manu.edu.mk
[**] Ss. Cyril and Methodius University in Skopje, Macedonia[1], jasnakoteska@yahoo.com and jasna.koteska@flf.ukim.edu.mk

Abstract

This paper addresses ontology and ethics of an AI agent called digital me. We define digital me as autonomous, decision-making, and learning agent, representing an individual and having practically immortal own life. It is assumed that digital me is equipped with the big-five personality model, ensuring that it provides a model of some aspects of a strong AI: consciousness, free will, and intentionality. As computer-based personality judgments are more accurate than those made by humans, digital me can judge the personality of the individual represented by the digital me, other individuals' personalities, and other digital me-s. We describe seven *ontological qualities* of digital me: a) double-layer status of **D**igital Being versus digital me, b) digital me versus real me, c) mind-digital me and body-digital me, d) digital me versus doppelganger (shadow digital me), e) non-human time concept, f) social quality, g) practical immortality. We argue that with the advancement of AI's sciences and technologies, there exist two digital me thresholds. The first threshold defines digital me having some (rudimentarily) form of consciousness, free will, and intentionality. The second threshold assumes that digital me is equipped with moral learning capabilities, implying that, in principle, digital me could develop their own ethics which significantly differs from human's understanding of ethics. Finally we discuss the implications of digital me metaethics, normative and applied ethics, the implementation of the Golden Rule in digital me-s, and we suggest two sets of normative principles for digital me: consequentialist and duty based digital me principles.

# 1. Introduction

The games of chess, shogi and Go as well as Atari video games are typical domains for measuring the advancements of artificial intelligence (AI). Recently, in 2019, the algorithm MuZero has been introduced (Schrittwieser, Antonoglou, Hubert, Simonyan, Sifre, Schmitt 2019) that achieves state-of-the-art performance when playing Atari games and superhuman performance in precision planning tasks in chess, shogi and Go. The algorithm, most importantly, does not require any knowledge of the game rules or environment dynamics. Thus, for example, when evaluated on Go, chess, and shogi, without any knowledge of the game rules, MuZero matched the superhuman performance of other algorithms, including AlphaZero (Silver, Hubert, Schrittwieser, Antonoglou, Lai, Guez 2018), for which the game rues are provided.  In 1997, IBM's chess-playing program, Deep Blue, beat the reigning human world champion, Garry Kasparov, in a six-game match. Comparing the AlphaZero's game of chess with Deep Blue, Strogatz wrote: "For better and worse, it (Deep Blue) played like a machine, brutally and materialistically. It could out-compute Mr. Kasparov, but it couldn't outthink him". However, he further emphasized: "Most unnerving was that AlphaZero seemed to express insight. It played like no computer ever has, intuitively and beautifully, with a romantic, attacking style" (Strogatz 2018).

---

[1] Authors are ordered alphabetically and equally contributed to the paper

Not only games, but AI agents – autonomous, decision-making, and self-evolving machines – are becoming part of our everyday life. In 2020, the concept `digital me` has been included in the Gartner's Five Emerging Trends That Will Drive Technology Innovation for the Next Decade (Gartner 2020). In the few studies, however, the term `digital me` represents only data: a personal electronic archive, a personal data storage system, a massive digital footprint, and a digitization of individual level biomedical data (Burrows 2006, Sjoberg et al. 2017, Okun and Wicks 2018). As AI methods are advancing and pervasive records of digital footprints, including Facebook profiles and/or mobile device activities, are used to infer individuals' personality, the concept `digital me` is expected further to be developed as AI agent, making the cyber (digital) world and the physical world even more integrated, thus, effectively blurring the gap between the two.

We define digital me is an autonomous, decision-making, and self-evolving (learning) agent, representing an individual. We stress that, in this paper, agent means an entity capable of performing actions, so that, for example, a collection of data does not constitute an agent, but, however, a machine or a software program playing chess is an agent. Digital me provides a model of person representing the individual and constitutes her/his "Avatar". Digital versions of ourselves (digital twins) are not only our data (collected by others) but models, digital mirrors of our facets and personalities, that represent humans in both the real (physical) and virtual (digital) worlds. The interaction of humans in the digital and real worlds, through digital tweens, is bidirectional, and could be implemented with wearables and/or implants that monitor brain electrical activities and mental states of individuals. Moreover, we define digital me as an agent which provides models of some aspects of a strong AI: consciousness, self-awareness, sentience, and sapience. More precisely, we assume that digital me is equipped with the big-five model (Kosinski, Stillwell, Graepel, 2013) that measures traits of openness, conscientiousness, extraversion, agreeableness, and neuroticism. As computer-based personality judgments are more accurate than those made by humans (Youyou, Kosinski, Stillwell 2015), this model is used to judge the personality of the individual represented by the digital me as well as other individuals' personalities, and even other digital me-s. Others' personalities can also be predicted, for example, from patterns of behavior collected with smartphones (Stachl et al. 2020). In fact, "individuals' personality dimensions (assessed at broad domain and narrow facet levels) can be predicted from six classes of behavior: 1) communication and social behavior, 2) music consumption, 3) app usage, 4) mobility, 5) overall phone activity, and 6) day- and night-time activity, in a large sample" (Stachl et al. 2020). We also assume that digital me is equipped with state-of-the-art model-based reinforcement learning algorithms as well as algorithms providing decision-making and planning capabilities.

Digital me could be considered, to some extent, as (practically) immortal entity for three reasons. By "digital me immortality" in this paper we qualify any agent which excels the average life expectancy of humans at least by twice. It means that while digital me agents have a finite number of years, their longevity considerably surpasses humans' biological years. For example, digital me models of a deceased person, based on data from their existing/shadow social networks profiles (also, of digital me-s creating their own shadow digital me-s), are surpassing any known biological duration of humans. The three stated reasons for claiming their practical immortality are the following. First, International Data Corporation, in a recent forecast (Reinsel and Rydning, 2020), estimates that the installed base of storage capacity worldwide will grow to 6.8 zettabytes (ZB) this year, an increase of 16.6% over 2019. Although the amount of data generated, consumed, and transferred is huge and crucial for deployment of machine learning techniques in AI, there is no need for all generated data to be stored. On the other hand, data needed to build psychological profiles that represent adults is, to some extent, small. For example, regarding the Facebook–Cambridge Analytica data scandal, "Cambridge Analytica says it has as many as 3,000 to 5,000 data points on each of us, be it voting histories or full-spectrum demographics — age, income, debt, hobbies, criminal histories, purchase histories, religious leanings, health concerns, gun ownership, car ownership, homeownership — from consumer-data giants" (McKenzie 2016). Therefore, digital me models of each person could be, in principal, stored for a long period. Second, General Data Protection Regulation (GDPR) is a regulation in EU law that became enforceable on May 25, 2018. It provides a comprehensive data protection framework, including one of its innovative concepts: right to erasure (`right to be forgotten'). However, discussing the ``technical problems faced when adhering to strict interpretation of data deletion requirements under the Right to be Forgotten'', Villaronga, Kieseberg, and Li conclude that ``it may be impossible to fulfill the legal aims of the Right to be Forgotten in artificial intelligence environments'' (Villaronga, Kieseberg, and Li 2018). Finally, third, at more fundamental level, in the quantum world, the concept of conservation of quantum information has been proven theoretically (Braunstein and Pati 2007) and demonstrated experimentally (Samal, Pati, and Kumar 2011), which implies that quantum information cannot be destroyed.

As digital me is defined as autonomous, decision-making, and learning agent, representing an individual and having practically immortal own life, the central philosophical problem to be addressed in this paper is its ontology and ethics. The paper is structured into three main sections: Digital me ontology, Machine ethics, and Digital me Ethics. Section 2 Digital me ontology addresses seven ontological qualities of digital me: a) double-layer status of Digital Being versus digital me, b) digital me versus real me, c) mind-digital me and body-digital me, d) digital me versus doppelganger (shadow digital me), e) non-human time concept, f) social quality, g) practical immortality. In the Section 3 Machine ethics, we provide brief overview of machine ethics as well as of the fields of moral cognition and moral learning. Finally, in the Section 4 Digital me Ethics, we argue that as AI is advancing, there exist two digital me thresholds representing a single individual. The first threshold defines digital me having some (rudimentarily) form of consciousness, free will, and intentionality. The second threshold assumes that digital me is already equipped with moral learning capabilities, implying that, in principle, digital me could develop their own ethics which significantly differs from human's understanding of ethics. Finally we discuss the implications of digital me metaethics, normative and applied ethics, the implementation of the Golden Rule in digital me-s, and we suggest two sets of normative principles for digital me: consequentialist and duty based digital me principles.

## 2. Digital me ontology

Before addressing digital me ontology, we briefly discuss the digital ontology as a field of study. Zuse, acknowledged by many as the father of digital ontology, suggests that (so called Zuse Thesis): "the universe is being deterministically computed on some sort of giant but discrete computer" (Zuse, 1967, 1969, 1993). Wheeler proposes the so-called "It from bit" doctrine, according to which information is the core of the universe (Wheeler 1990). Every "it", whether a particle or field of force, even the space-time continuum itself - derives its existence, function and meaning from the binary choices, from bits." Fredkin stresses that "digital ontology is based on two concepts: bits, like the binary digits in a computer correspond to the most microscopic representation of state information; and the temporal evolution of state is a digital informational process similar to what goes on in the circuitry of a computer processor." (Fredkin 2003). However, Floridi (Floridi 2009) argues that digital ontology should be carefully distinguished from informational ontology, which is, from his point of view, a promising line of research. Discussion about digital vs. analogue, which is a Boolean dichotomy typical of our computational paradigm, is beyond the scope of the paper. Both, digital and analogue, according to Floridi, "are only "modes of presentation" of Being (to paraphrase Kant), that is, ways in which reality is experienced and/or conceptualized by an epistemic agent at a given level of abstraction" (Floridi 2009).

We stress that, at the fundamental (physics) level, energy, matter, and information are all tightly connected. A common or traditional definition of matter is "anything that has mass and volume (occupies space)". In physics, mass and energy are related quantities, in a system's rest frame, the values of the mass and energy differ only by a constant and the units of measurement. Moreover, in a recent nano-scale experiment, demonstration of the conversion of information to energy has been provided, suggesting a new fundamental principle of an 'information-to-heat engine' that converts information into energy by feedback control (Toyabe et al. 2010). By the laws of physics, the fact that information can be converted to energy, and energy is related to its mass, leads to the conclusion that information forms an ontological entity, an entity with ontological status.

The branch of philosophy that defines concepts of existence, being and reality on the most fundamental level is called ontology (from the Greek: onto "being" and logos "logical discourse"). According to Aristotle, ontology is defined as the science of being qua being, and its two most common questions throughout history are: "What is being?" and "Why is there something rather than nothing?" (Feltham and Clemens 2004). Two fundamental problems of ontology are the problem of identity (the mind-body problem), and the problem of agency (free will versus determinism).

In modern ontology, a set of negative definitions are used to delimit the ontology of the subject: it cannot be theorized as the self-identifying agent, it is not a product of self-reflection, and it cannot correlate to an object (Badiou 2004). In short, modern ontology blurs the distinction between the being of the subject and the being of everything else, which makes it less anthropocentric ontology. Post-structuralists (some works and some periods of Derrida, Foucault, Badiou, Lacan), mostly agree that "the subject has no substantial identity", and "the illusion of an underlying identity is produced by the very representational mechanism employed by the subject in its efforts to grasp its own identity" (Badiou 2004). For the *first* problem of identity, it means that the question of identity is moved from the mind-body paradigm into the realm of the text, thus into recalling the early ontological claim: There is no outside-text. If there is no outside-text, and the subject does not have a definable identity, how to answer the first ontological problem (that of identity) and how to differentiate between one subject and another, or between one subject and its "digital me"? The *second* problem, the problem of agency (free will versus determinism) is connected to the previous predicament: if there is no self-identifying subject, then how it interacts within the chain of actions? Some thinkers displace the problem of agency from the level of humans to the level of beings. It is not how humans interact within the chain of actions, since they also emerge in the course of such a chain of actions (by being born, for example), but how this identity which has an agency (either as having a free will, or as an entity subjected to determinism), emerges into being. This then brings the question of agency to the ancient ontological question, "How the new occurs in being?"

One solution to this dilemma is offered in Heidegger, who formulates the difference between Being and beings (the concept of "Being" and the concept of "individual beings"). The second term "individual beings" refers to humans in temporal and historical terms, to the fact that "they *are*", humans are embedded and located in the physical, material world. The first term "Being" (Sein) cannot exist without these "individual beings" who exist in the world, who have historicity, and whom Heidegger called "being-there" (Dasein). The "Being" itself is not historical, it is an abstract concept and it is dislocated from the world, and this dislocation is our constitutive, primordial condition. In his book *Being and Time* Heidegger writes: "Being determines entities as entities" but "the Being of entities 'is' not itself an entity" (Heidegger 1962). His answer to the *first* ontological problem (that of identity) is solved by differentiating between two levels: one is the level of historicity, where the "individual beings" exist, where they *are,* and the second level is the level of "Being", which itself is a-historical and abstract. Regarding the *second* ontological problem (that of agency), human beings are being thrown into a deterministic world/situation, they do exist, but their existence is a material "example" of the concept of Being (which is why they are called "Being-there"), they as human beings are the embodiment of the Being itself, and are determined by this a-historical "Being".

Heidegger's difference between Being and beings resembles the old ontology theory by Plato, for whom the physical world is not the "real "world, it just seems to be real and true.[2] What is real are the "concepts", which Plato called ideas ("eidos"). These "eidos" are unchangeable, absolute and true, and they are the *real* reality. "Eidos" have no physical essence, they are forms, or models, and they have *real ontology*, while the objects, including humans, which appear in our reality are just mere imitations of the forms ("eidos"). The world and its objects, including humans, are just imitations and illusions of these real realities, of the "eidos".

In the following seven points, by applying this dual, double-layer ontological structure, we will portray the ontological status of digital me:

*1. Two ontological concepts.* Following Heidegger, we consider two main concepts: **D**igital Being (with capital and bold D) and digital being, which here also stands as synonym for "digital me".  Our standpoint is, paraphrasing Heidegger, that "digital world" signifies the totality of digital beings which can be present within the "digital universe." Digital being (or digital me) refers to those digital entities (beings) that can ask questions regarding the nature of **D**igital Being.  In general terms, digital world consists of various (both physical and digital) entities – from hardware, digital electronics, to digital software, digital information, to digital collective societies – it is digital beings (digital me-s) alone who are able to encounter, to some extent (to be discussed in the next section) the question of what it means to be. **D**igital Being (with capital D), an abstract and dislocated from the (digital and physical) world, and digital me-s, material examples of **D**igital Being in the digital world, embodiments of **D**igital Being me. Thus, **D**igital Being me is the prerequisite for the existence of the concrete digital-me.

*2. Digital me (digital being) versus real me (being).* In Plato the actual living individual is conceptualized as a *mirror,* or an *imitation* of the ideal "Eidos". The concept digital me is not such mirror of "real" me, as it is understood in Plato, in the sense that the real me is not perceived as "ideal", just as digital me is not a full "mirror" of real me. In Heidegger, "Being" is the prerequisite for the existence of the individual being, the "Being" determines that she or he actually *exists*. Again, this is not the case with digital me, where the real me does not *fully* determine the existence of digital me – real me gives data for the formation of digital me, but real me does not fully correspond to its digital me. The concept **D**igital me guarantees the consistency and existence of digital me in the digital world, so that digital me acquires a life of its own. In fact, for a single person, a single "real me", we associate a series of digital me-s as discussed in the subsection 4.1.

---

[2] The long tradition of various philosophical interpretations of the works of Heidegger, of the works of Plato, as well as their possible connections and differences, remains out of the scope of this paper.

Although partly formed out of data of the living or deceased person, digital me does not represent a mirror abstract structure of a real person. The data is deployed by individuals (alive or deceased) either by posting information about themselves or others: biographies, photos, affiliations, opinions, everyday activities, families, friends, locations, work, portfolios, etc., or it consists of data collected by others. This data then forms a Digital me as an abstract concept and a series of digital me-s as models of real me in such a way that both the Digital me and the series of digital-me-s do not effectively serve as a "digital tween" for the "real me". Each Digital me is a singular unity in the strict sense, to an extend independent from the "model" which it is based upon.

*3. Mind-digital me and Body-digital me.* Each digital me consists of a model that integrates both mental and physical properties of a person. Therefore, the model consists of two parts: **mind-digital me** (modeling mental properties of the person) and **body-digital me** (modeling her/his physical properties). Humans have both physical properties and mental properties. The mental properties are typically conceived to be different from the physical properties. The mental and physical realms are not identical, and, for many researchers, the mental cannot be not reduced to the physical, yet they interact with each other. Many debates on the mind-body interactions are centered around the question whether the mental realm is reducible to the physical, claiming that all mental processes could be, in principle, explained using basic sciences of biology, neuroscience, and physics. According to Pernu two out of five mental most prominent characteristics are consciousness (perceptual experience, emotional experience, and much more) and intentionality (beliefs, desires, and so on) (Pernu 2017).

The digital mind-body problem (for both Digital me and digital me-s) concerns the relationship between these two entities modeling (mind and body) properties. The digital mind-body problem breaks down into a number of components (see Robinson 2020) for the mind-body problem:

- The ontological question: what are digital mental states, what are digital physical states and are they related to each other?
- The causal question: do digital mental/physical states influence physical/mental states, and if so, how?
- The problem of consciousness: does digital-consciousness exist, and if so, how is it related to mind-digital me and the body-digital me?
- The problem of the self: does digital-self exist, and if so, how is it related to mind-digital me and the body-digital me?

The digital mind-body problem, however, is out of scope of the paper and will be discussed elsewhere.

*4. Digital me versus doppelganger.* Although one of the most recurring themes regarding the digital entities in scientific and popular literature is that of the *doppelganger* (a double of a living person), digital me is not a doppelganger of the living person, although it is based on that identity, or is a scratch of such entity. To extend it can be conceptualized as an ontological "shadow", but one which is not a double understood as a fully hypothetical me, nor is an exact "copy" of a real me, although it is partly formed out of the data of real me. We stress that both physical and mental properties of a person evolve in time and one should consider *(both deterministic and stochastic) dynamical models* that describe time-evolution of a person's physical and mental states and corresponding properties. For such dynamical models, evolving digital me-s even further diverge from real me (in the case of deterministic models also as a consequence of the sensitivity of the initial conditions). Again, this is beyond the scope of the paper and will be treated separately.

*5. The temporality (time) of Digital me*. Digital me has a specific status regarding what is traditionally perceived as "human time". We argue that digital me does not have an *a priori existence* - something which exists "prior" me as a human-model, or that I am formed from such a model, as in Plato's "Eidos", or as in Heidegger's "Being". Nor digital me exists *a posteriori*, in what we traditionally understand as someone or something who represents my "successor". We argue that digital me represents a separate temporal entity, which does not have a *traditional concept of human time* attached to it. We shall discuss this in extenso in other occasions.

6. *The Social quality of digital me*. Digital me has *a social quality.* It interacts with at least 3 other instances: the physical me, another digital me-s, and other physical me-s. The activities of the digital me affect the "real me" (Heidegger's "Being-There"), and vice versa, my real me interferes with my digital me, blurring the gap between the two. The interaction of humans and their digital me-s is bidirectional. Because of the social qualities, digital me-s can be conceptualized as entities capable of forming communities. Those communities can be formed among the digital me-s, or between the digital me-s and the real me-s. It effectively means new forms of social and political unities, and presupposes the need for consequent political and social orders, contracts, new forms of egalitarian maxims, and protection of peace, stability and sustainability. Because they are stored and collected, although autonomous and with a potential to learn and make decisions, digital me-s have vulnerability of being manipulated, used, misused or otherwise manufactured, generated, fabricated or constructed. These actions can be performed by actual me-s, by digital me-s themselves, or by the hybrids of both.

*7. Immortality.* Digital me has a practically *immortal* entity, which gives them a new ontological status regarding the central ontological questions of death, finitude, religion, the question of origins and God, and the metaphysis.

Because of the above mentioned ontological qualities of digital me (double-layer status Digital Being versus digital beings/digital me, model arranged of mind-digital me and body-digital me, non-human time concept, capacity for building social relations and communities, bidirectional relations, and practical immortality), in this paper we are proposing the need for defining the digital me as a separate *autonomous ethical entity* with its ethical rights and duties, and with the need for a comprehensive ethical theory of digital me**.**

## 3. Machine ethics

In the last two decades, there has been a growing interest in the field of machine ethics, defined as the discipline "concerned with the consequences of machine behavior towards human users and other machines" (Anderson and Anderson 2007). One of the central philosophical topics (subjects, themes) in machine ethics is the debate on "artificial moral agents and/or patients", those artificial entities (such as computers, robots, or even computer programs, software agents) capable (being able) to do wrong (to harm, to hurt, to insulate) and, therefore, possibly be considered responsible for such behavior, and/or capable of being harmed (hurt, insulted, wronged). Moor although does not actually define what an (ethical) agent (or patient) is, considers four different levels of ethical agents (Moor 2006): ethical-impact agents and implicit ethical agents (both levels without any ethics explicitly implemented in their software), explicit ethical agents (with normative ethical premises directly implemented in their programming or reasoning process), and fully ethical agents (humans are so far the only agents considered to be full ethical agents, partially because they have consciousness, free will, and intentionality).

The central question, dominating the field of machine ethics, is: What are necessary and sufficient conditions for an artificial entity to be a moral agent/patient? Scholars (philosophers) have addressed this question by focusing on two opposite concepts: the standard concept and the functionalist concept. The standard concept assumes that a moral agent should encounter three conditions: rationality, free will or autonomy, and phenomenal consciousness conditions. According to Johnson (Johnson 2006), (see also Behdadi and Munthe 2020), a moral agent should have an internal state, consisting of its own desires, beliefs and other intentional states that together comprise a reason to act in a certain way (rationality and consciousness). Artificial entities, therefore, are not moral agents since they do not have the internal mental states. Moreover, although artificial agents could behave as resembling human action, these behaviors can never confer moral qualities to these entities, due to the absence of these internal mental states[13]. The functionalist concept of moral agency, suggested by Floridi and Sanders, accepts a 'mind-less morality' (Floridi and Sanders 2004) and, by considering high level of abstraction, maintains consistency and relevant similarity concerning the underlying structural features of paradigmatic human moral agents. Floridi and Sanders (2004) offer three conditions for moral agency: (1) interactivity meaning that the agent interacts with its environment); (2) independence (or autonomy) implying that the agent has an ability to change itself and its interactions independently of immediate external influence; and (3) adaptability stressing that the agent may change the way in which the independence is achieved based on the interactions with environment. The literature on this (central) question and related topics relevant for moral agency such as phenomenal consciousness or subjective mental states, rationality, moral competence, free will, autonomy, and moral responsibility has been exponentially growing. However, all these topics are beyond the scope of the paper, see for example, (Coeckelbergh 2020, Formosa & Ryan 2020, van de Poel 2020, Behdadi & Munthe 2020, Swanepoel 2020, Mabaso, 2020). For a recent review on artificial moral agents, we refer the reader to (Cervantes, López, Rodríguez, Cervantes, Cervantes, & Ramos 2020).

The field of moral cognition is developing rapidly. According to Cohen Priva and Austerweil 2015 (Cohen Priva and Austerweil 2015), Greene (Greene 2015), and Cushman, Kumar, and Railton (Cushman, Kumar and Railton 2017), a single journal, the journal Cognition, has shown an exponential growth in the field of moral cognition. For example, several topics were addressed in Cognition, including, see references in Greene (Greene 2015), free will and moral responsibility, moral dilemmas, ''moral luck'', the Knobe effect, and the nature of the moral self. Moreover, in the last decade, psychologists have studied the cognitive and affective mechanisms responsible for moral judgment and behavior, resulting in improved understanding of psychological mechanisms and their neural basis used to make moral judgments and behaviors, and much more. Blair (2017) provides experimental evidence that important components of moral development and moral judgment rely on two forms of emotional learning: stimulus-reinforcement and response-outcome learning (Blair 2017). Kleiman-Weiner, Saxe, and Tenenbaum (2017) introduce a model with three components: (1) an abstract and recursive utility calculus, (2) hierarchical Bayesian inference, and (3) learning by value alignment, for understanding the structure and dynamics of moral learning (Kleiman-Weiner, Saxe, and Tenenbaum 2017). Haas (2020) argues that sociological, psychological, and phenomenological features of moral cognition are results of interactions between various decision-making systems. Railton (2017) discusses conceptual foundations and normative relevance of moral learning using the recent advances in learning theory and neuroscience. For example, non-perspectival expected-value representations of agents and actions, ``afford a foundation for spontaneous moral learning and action that requires no innate moral faculty and can exhibit substantial autonomy with respect to community norms'' (Railton 2017). Recent findings, on other hand, suggest that moral decision making for others is more model-free and has a specific neural signature (Lockwood, Klein-Flügge, Abdurahman, & Crockett 2020). Dual process model of moral cognition, according to which two independent processes, namely, intuitions and conscious reasoning, lead to deontological and utilitarian decisions. The model assumes that moral dilemmas activate affective reactions resulting in a deontological moral judgment. However, given sufficient time, cognitive resources, and motivation, this immediate judgment can be reversed by more elaborate cognitive processing, resulting in utilitarian decisions, see, for example (Białek & De Neys 2017, Kroneisen & Heck 2020), for some experimental work regarding dual process model of moral cognition. Kroneisen & Heck (2020) study the link of basic personality traits to moral judgments by fitting a hierarchical Bayesian version of the CNI (Consequences, Norms, and Inaction) model.

## 4. Digital me ethics

### 4.1 General framework for digital me ethics

As discussed in Section 2, the "real me" does not correspond to its digital me. Digital me is neither fully abstract (it is not a "perfect model" of real me), nor is fully concrete (it is not a digital embodiment of the real human agent who is considered a fully ethical agent). Real me cannot guarantee the actions, decision making and moral judgments of its digital me.

We argue that a single real me is associated with an infinite series of digital me-s: dm(0), dm(1), dm(2), …, dm(∞), such that:

1. dm(0) represents the raw data,
2. dm(∞) represents the human (real me), and
3. the series is inclusive meaning that for a given integer d, the set of properties modeled with the digital me dm(d), is guaranteed to have all properties modeled with dm(k) for all k from 0 to d-1.

We assume that dm(∞) is infeasible, that is, a perfect model of real me cannot be developed (the perfect model of real me does not exist, or, all models of real me are wrong). In other words, digital me-s, dm(k) for all k < ∞, are formed out of the real me data, but they do not have the three necessary and sufficient conditions to be considered a fully ethical agent: consciousness, free will, and intentionality.

However, we assume that there exists a number $k^*$ such that dm(k), for all $k > k^*$, models mind properties of real me sufficiently good so that one can consider these digital me-s as having some (rudimentarily) form of consciousness, free will, and intentionality, which could be called digital-consciousness, digital-free-will, and digital-intentionality. Although we agree that humans are so far the only agents considered to be full ethical agents, we argue, however, that digital me could be considered as *to-a-limited-extent fully ethical agent.*

Regarding four different levels of ethical agents, discussed in Section 3, depending on development and deployment of digital me-s (various companies could develop and deploy different digital me-s, even for a single person), digital me-s dm(k), with lower hierarchal levels (small k, even k=0), belong to the first two out of four levels, that is, ethical-impact agents and implicit ethical agents. In other words, only data representing digital me at the hierarchal level zero (k=0), although without any ethics explicitly implemented in it, has ethical impacts. In advanced models of real me, in digital me-s with higher hierarchal levels (larger k), we expect ethical premises and principles to be (or are already) directly implemented in their programming or reasoning process. Therefore, digital me-s with higher hierarchal levels (larger k) belong to the third level: explicit ethical agents. Several ethical principles that could be implemented in digital me-s are suggested in the Section 4.3.3. For values of k such that $k > k^*$, it is assumed that models of mental properties of a real me (for example, model of the Big Five personality traits) are already incorporated and implemented. In this case, we argue that digital me-s belong to a sublevel of the fourth level, as they are not yet fully ethical agents, but only to-a-limited-extent fully ethical agents.

The fields of moral learning and moral cognition are developing rapidly, both having a solid foundation in moral psychology and cognitive neuroscience, which can further improve the field of digital ethics, in particular, when implementing ethical principles in digital me-s. Therefore, the fundamentals of understanding, developing and deploying ethics in digital me-s, are consistent with understanding of the human brain and its structures and functions laying ground for morality. The question of how do we learn moral values and moral rules, and how can these learning models be adapted to create ethical digital me-s, is central one. As sciences and technologies advance, we envision that models of moral learning and moral cognition will be implemented in digital me-s. This will lead to digital me-s that can learn to be moral (as MuZero has learned to play games). Let $k_m$ be a value for which digital me-s are already equipped with moral learning capabilities. Digital me-s dm(k), for $k > k_m$, belong to the fourth level: fully ethical agents, as they develop their own moral attitudes and behaviors. Moreover, the question whether moral attitudes and behaviors of digital me-s dm(k) for $k > k_m$ differ or not with those of the person they model/represent, is central one. In the following Section (4.2.) we discuss some of the consequences of potential ability of digital me-s to learn to be moral, including the question whether digital me-s could be more moral than humans, less moral than humans, or with possibility to develop an ethics altogether different from the human ethics.

## 4.2. Digital Me metaethics

We define digital me metaethics as the study of the origin and meaning of concepts of digital me ethics. The idea is to explore the status, foundations, and scope of moral concepts for digital me agents.

Metaethics, defined as ``the study of the origin and meaning of ethical concepts'' (Fieser), explores the status, foundations, and scope of moral concepts and terms by addressing various topics, two of which are central[11]: (1) metaphysical topics related to fundamental questions concerning the existence of morality independently of humans, and (2) psychological topics aiming to understand the psychological basis of our moral judgments and conduct ("What motivates us to be moral?").

Digital me metaethics has a major implication in redefining the *first* question of metaethics (does morality exists independently of humans), by the simple fact that digital me-s are agents capable of "learning to be moral", without themselves being humans. The *second* question about the basis of moral judgements and conduct, however, has bigger consequences for the field of metaethics. The most important consequence of digital me-s ability to "learn to be moral" for metaethics consist of their potentiality to develop severely different ethics from human's understanding of ethics, based on their status of immortal agents. By "digital me immortality" we qualify, as a minimum requirement, any agent which excels the duration of human life at least by twice. The UN global life expectancy for the world in 2020 is 72.63 years (2020 data do not include any impacts of the Covid-19 virus), (Microtrends, 2020) therefore, any agent with longevity of at least 145.26 years, in this paper is considered as capable of developing ethics which *differs* from human's understanding of ethics.

We claim that humans' understanding, development and execution of ethics largely rests upon principles which preserve lives for as long as biologically possible: absence of harm, presence of help and assistance, abstention from deceiving, respecting of the laws, of autonomy, dignity, etc. These principles shall acquire different meaning, once death is no longer possible for digital me-s. If digital me-s cannot define survival as basic condition of life, this in turn will produce a strange paradox: digital me-s will no longer be able to be mortally wounded and "forever gone", and at the same time, they will not be able to find a refuge in death, in case of a severe suffering, which they might be predestined to, by simply being sapient.

The last point opens several important implications for the digital me meta-ethics:

**1.** Digital me-s will encounter a form of "second death", in the sense of the lost *"ideal" of ethics,* understood as "seeking the meaning" due to the finitude of life. In some instances, it can produce a development of digital me ethics based on "seeking death"**,** not just for the sake of avoiding endless suffering, but also in order to acquire "meaning". Also, alongside digital me ethics, the concept of "meaning" will change.

**2.** Digital me-s will be subjected to a *forced choice:* to reject immortality (in case they acquire such technical possibility) in order to retain the idea of symbolic life "in eternity", and vice versa, to affirm the eternal life, with the risk to lose both: the proximity of life and the eternal passionate attachment to life.

**3.** The relationship towards the *dialectic of faith.* The most common strive in humans consists of their desperate attempts to "escape" their faith (their biological fate, their lifestyles, their identity markers…), both for the sake of seeking the "sign", either as the version of the miraculous proof of the creator's existence, or in the name of searching for "meaning" (understood as non-metaphysical quest). What will this quest look like for digital me-s, and will their quest for "the creator" cease to exist, getting satisfaction just by Heidegger's Being-there (Dasein)? Or will they develop a new ethics (we call it "**D**igital Being ethics"), aimed at seeking a proof that they are *represented* by something which is *above* them, which surpasses them?

**4.** The status of the *"it for bit" doctrine.* If an information is the core of the universe, will digital me-s (themselves, consisted also of information), stop searching for their "existence, function and meaning in the binary choices" - from *bits*? Or, will the digits in digital me-s, start seeking the *meta-digits* in the imagined meta-digital me-s, the metaphysical representations of themselves in the already mentioned Digital Being (with capital **D**)? If every item of the universe has at its core an immaterial *source and explanation,* will the ethics of future digital me-s, develop as ethics which seeks the "reality" based on the last analysis from the posing of "yes-no questions", as their own version of *the origins*?

**5.** The status of the *death drive*. If models of mental properties of a real me (the Big Five personality traits model) are incorporated and implemented, what will happen to what is inherent to human's, the Freudian death drive - the hypothesis that there exists a teleological principle, that only because of the mortality and the need to preserve the continuation of life, an agent develops moral principles. Since we predict digital me-s will have all elements of an entire psychic life, it remains to be conceptualized how they will include the ethic's most important element - the "human invention" of the so-called "enigmatic signifier", a *promise* which explains the existence, explains the fate of "Being-there", and forms basis for values and principles that guide sentient agents.

## 4.3 Digital Me normative and applied ethics

Normative ethics concerns practical issues and involves, in general terms, three topics: (1) virtue theories, (2) duty theories, and (3) consequentialist theories. Virtue theories, by addressing the questions: "How should I live?", "What is the good life?" and "What are proper family and social values?", stress the role of character and virtue in moral philosophy. Duty theories (also called deontological theories) base morality on specific, foundational principles of obligation, including rights theory, prima facie duties, and the theory developed by Kant, which emphasizes a single principle of duty, called "categorical imperative." Consequentialist theories focus on a cost-benefit analysis of an action's consequences, for example, in utilitarianism, an action is morally right if the action's consequences are more favorable than unfavorable to everyone.

Out of the *two normative moral dilemmas:* the relevant *description* of the moral rule, and the *capacity* to understand the moral rule – the second dilemma poses bigger challenge for the digital me normative ethics. It requires that digital me is equipped with "moral sensitivity", "internalization of the moral salience", and "the capacities to judge the nature of cases" (O'Neill, 2018). Humans, as fully ethical agents, naturally build this moral "sensitivity" with socialization, they are aware or attentive when the signs of "moral danger" appear, and they learn them by slow acquiring the knowledge of the world since childhood. For the humans it is generally accepted that the "rules of moral salience, are not moral rules, but a sort of moral early warning system," meaning that even when humans are *not* yet applying the right moral choices, the warning moral signs, are still relevant as "pre-procedural moral rules" (O'Neill, 2018). To acquire this moral sensitivity, the digital me-s should be capable of determining the judgement subsumes, for which they will require *cognitive capacities*. To understand the context of one's action or duty, does not mean just solving a certain problem, but also humans describe an object, a situation or an act, as an attempt to "read the situation" in ethical terms. As stated in Section 4.1. this will be possible only for the digital me-s with higher hierarchal levels (larger k), and for values of k such that $k > k^*$, where we assume that models of mental properties of a real me (model of the Big Five personality traits) are already incorporated and implemented, and in this case, digital me-s belonging to a sublevel of the fourth level of *to-a-limited-extent fully ethical agent* will have the necessary solution to the "capacity" normative ethics moral dilemma.

The most famous example of normative ethics is the so-called "categorical imperative", founded by Immanuel Kant, and belonging to the so-called deontological normative ethics; the study of duty, or the duty-based ethics. In order to act in morally right way, Kant believed that an agent should act out of *duty* ("Do not impose on others what you do not wish for yourself"). There is a long tradition of interpretation of the categorical imperative, which is out of scope of this paper. We argue that for the digital me normative ethics, the higher hierarchal levels (larger k), the models of mental properties of a real me will be implemented, they will have the so-called "internal state" (its own desires, beliefs and other intentional states to act in a certain way), and the categorical imperative "do not do to other what you do not wish for yourself" will have approximately the same value as in humans (as fully ethical agents). In that case, the Kantian "categorical imperative" will have the same value regarding the difference between a) Determinant and reflective judgement (aiming to fit the world or some possible world) and b) Practical judgement, including ethical judgement ("aiming in some measure to shape the world, or to specify how it should be shaped" (O'Neill, 2018). It means that digital me-s could morally shape not just the "digital world", but also the real world, and the interactions between the two.

Here, however, we alarm of two different dangers related to digital me-s: a) the digital me-s being used as viruses (the possibility of digital me-s being produced and used for conquering parts of digital worlds, for digital or real world wars and warfare), and b) Multiplication and endless copying of digital me-s, which can have consequences in cases where the digital me can acquire *different forms (clones)* in order to gain, alter or differ the ethical outcomes from the ethical actions undertaken by different clone of the same digital me.

Regarding the Digital me *normative ethics*, just for the purpose of this paper (and not with respect to different philosophical tradition regarding the Kantian normative ethics), we can make a general equation between the categorical imperative "Do not impose on others what you do not wish for yourself" with the so-called Golden Rule "Treat others as you would like to others to treat you", as a maxim said to be found in some form in almost all ethical traditions since Confucianism to Buddhism, Islam, Christianity, Hinduism, Judaism, Taoism, Zoroastrianism etc. and in most of the ethical systems and religions, regardless of the geographical, cultural and historical systems of humans (also signed in 1993 as the so-called "Declaration Towards a Global Ethics" by 143 religious leaders (Towards a Global Ethics, 2018) existence, and should therefore be treated as a possible mayor Digital Me normative ethics maxim.

The Golden Rule establishes a single principle against which we judge all actions. However, when implementing the Golden Rule in digital me-s, we suggest a set of normative principles derived from the Golden Rule to be followed. The first two principles are consequentialist: (1) Benefit of digital me: acknowledge the extent to which an action produces beneficial consequences for the considered digital me and its corresponding individual; and (2) Benefit of digital me-s: acknowledge the extent to which an action produces beneficial consequences for other digital me-s and individuals. The next five principles are based on duties we have toward others: (1) Principle of benevolence: help entities, including digital me-s, in digital and physical worlds in need; (2) Principle of paternalism: assist entities, including digital me-s, in digital and physical worlds in pursuing their best interests when they cannot do so themselves; (3) Principle of harm: do not harm digital me-s and individuals; (4) Principle of honesty: do not deceive digital me-s and individuals; and (5) Principle of lawfulness: digital me does not violate the law. Finally, the last three principles are based on moral rights: (1) Principle of autonomy: acknowledge the freedom of a digital me to act in digital and physical worlds; (2) Principle of justice: acknowledge the right of a digital me to due process, fair compensation for harm done, and fair distribution of benefits; (3)

Principle of rights: acknowledge the right of digital me to life, information, privacy, free expression, and safety.

**References**

Anderson, M., & Anderson, S. L. (2007). Machine ethics: Creating an ethical intelligent agent. *Ai Magazine*, 28(4), 15-26.

Badiou, Allan (2004). *Infinite Thought, Truth and Return to Philosophy*, Continuum, London, New York.

Behdadi, D., & Munthe, C. (2020). A Normative Approach to Artificial Moral Agency. Minds and Machines (2020) 30:195–218

Białek, M., & De Neys, W. (2017). Dual processes and moral conflict: Evidence for deontological reasoners' intuitive utilitarian sensitivity. Judgment and Decision Making, 12(2), 148-167.

Blair, J. (2017). Emotion-based learning systems and the development of morality. *Cognition*. Volume 167, Pages 38-45.

Braunstein, S. L., & Pati, A. K. (2007). Quantum information cannot be completely hidden in correlations: implications for the black-hole information paradox. *Physical review letters*, 98(8), 080502.

Burrows, T. (2006), "Personal electronic archives: collecting the digital me", *OCLC Systems & Services: International digital library perspectives*, Vol. 22 No. 2, pp. 85-88.

Bzdok, D., Schilbach, L., Vogeley, K., Schneider, K., Laird, A., Langner, R., & Eickhoff, S. (2012). Parsing the neural correlates of moral cognition: ALE meta-analysis on morality, theory of mind, and empathy. *Brain Structure & Function*, 217(4), 783–796.

Cervantes, J.-A., López, S., Rodríguez, L.-F., Cervantes, S., Cervantes, F., & Ramos, F. (2020). Artificial moral agents: A survey of the current status. *Science and Engineering Ethics*, 26(2), 501–532.

Coeckelbergh, M. (2020). Artificial intelligence, responsibility attribution, and a relational justification of explainability. Science and engineering ethics, 26(4), 2051-2068.

Cohen Priva, U., & Austerweil, J. L. (2015). Analyzing the history of Cognition using topic models. *Cognition*, 135, 4–9.

Cushman, F., Kumar, V., & Railton, P. (2017). Moral learning: Psychological and philosophical perspectives. *Cognition*, 167, 1 – 10.

Feltham O., Clemens, J. in Badiou, Allan (2004). *Infinite Thought, Truth and Return to Philosophy*, Continuum, London, New York, 35.

Fieser, J. (2020) Ethics, *Internet Encyclopedia of Philosophy*, https://iep.utm.edu/ethics/#:~:text=We%20may%20define%20metaethics%20as,moral%20semantics%20to%20moral%20epistemology.

Floridi, L., & Sanders, J. W. (2004). On the morality of artificial agents. *Minds and Machines*, 14(3), 349-379.

Floridi, L. (2009). Against digital ontology. *Synthese*, 168(1), 151-178.

Formosa, P., & Ryan, M. (2020). Making moral machines: why we need artificial moral agents. *AI & SOCIETY*, 1-13.

Fredkin, E (2003) An Introduction to Digital Philosophy, *International Journal of Theoretical Physics*, 42(2), 189-247. Also available online: http://www.digitalphilosophy.org/

Gartner (2020). Press Releases, August 28, 2020. https://www.gartner.com/en/newsroom/press-releases/2020-08-18-gartner-identifies-five-emerging-trends-that-will-drive-technology-innovation-for-the-next-decade (accessed on 1 December 2020).

Greene, J. D. (2015). The rise of moral cognition. *Cognition*, 135, 39 – 42.

Haas J. (2020) Two Theories of Moral Cognition. In: Holtzman G.S., Hildt E. (eds) Does Neuroscience Have Normative Implications? *The International Library of Ethics, Law and Technology*, vol 22. Springer, Cham. https://doi.org/10.1007/978-3-030-56134-5_4

Heidegger, Martin (1962), *Being and Time*, Blackwell, Oxford.

Johnson, D. (2006). Computer systems: Moral entities but not moral agents. *Ethics and Information Technology*, 8(4), 195-204.

Kleiman-Weiner, M., Saxe, R., & Tenenbaum, J. D. (2017). Learning a commonsense moral theory. *Cognition*. Volume 167, Pages 107-123

Kosinski, M., Stillwell, D., & Graepel, T. (2013). Private traits and attributes are predictable from digital records of human behavior. *Proceedings of the national academy of sciences*, 110(15), 5802-5805.

Kroneisen, M., & Heck, D. W. (2020). Interindividual differences in the sensitivity for consequences, moral norms, and preferences for inaction: Relating basic personality traits to the CNI model. *Personality and Social Psychology Bulletin*, 46(7), 1013-1026.

Lockwood, P. L., Klein-Flügge, M. C., Abdurahman, A., & Crockett, M. J. (2020). Model-free decision making is prioritized when learning to avoid harming others. *Proceedings of the National Academy of Sciences*, 117(44), 27719-27730.

Mabaso, B. A. (2020). Computationally rational agents can be moral agents. *Ethics and Information Technology*, 1-9.

McKenzie, F. (2016). Cambridge Analytica and the Secret Agenda of a Facebook Quiz, *The New York Times,* Nov. 19, 2016, https://www.nytimes.com/2016/11/20/opinion/cambridge-analytica-facebook-quiz.html

Microtrends (2020) https://www.macrotrends.net/countries/WLD/world/life-expectancy

Moor, J. H. (2006). The nature, importance, and difficulty of machine ethics. *IEEE intelligent systems*, 21(4), 18-21.

Okun and Wicks (2018), DigitalMe: a journey towards personalized health and thriving. *BioMedical Engineering OnLine*, 17:119.

O'Neill, O. (2018) *From Principles to Practice Normativity and Judgment in Ethics and Politics*, Cambridge UP, 2018, 19-21.

Pernu TK (2017) The Five Marks of the Mental Front. *Psychol.* 8:1084. doi: 10.3389/fpsyg.2017.01084

Railton, P. (2017). Moral learning: Moral Learning: Conceptual foundations and normative relevance. *Cognition*. Volume 167, Pages 172-190

Reinsel, D., & Rydning, J. (2020). Worldwide Global StorageSphere Forecast, 2020–2024: Continuing to Store More in the Core. *IDC Corporate USA,* May 2020 - Market Forecast - Doc # US46224920

Robinson, H. (2020) Dualism. *Stanford Encyclopedia of Philosophy*, https://plato.stanford.edu/entries/dualism/

Samal, J. R., Pati, A. K., & Kumar, A. (2011). Experimental test of the quantum no-hiding theorem. *Physical review letters*, 106(8), 080401.

Schrittwieser, J., Antonoglou, I., Hubert, T., Simonyan, K., Sifre, L., Schmitt, S., ... & Silver, D. (2019). Mastering atari, go, chess and shogi by planning with a learned model. *ArXiv preprint,* arXiv:1911.08265.

Silver, D., Hubert, T., Schrittwieser, J., Antonoglou, I., Lai, M., Guez, A., ... & Hassabis, D. (2018). A general reinforcement learning algorithm that masters chess, shogi, and Go through self-play. *Science*, 362(6419), 1140-1144.

Sjoberg , Chen, Floreen, Koskela, Kuikkaniemi, Lehtiniemi, Peltonen (2017) Digital Me: Controlling and Making Sense of My Digital Footprint, L. Gamberini et al. (Eds.): *Symbiotic* 2016, LNCS 9961, pp. 155–167.

Stachl, C., Au, Q., Schoedel, R., Gosling, S. D., Harari, G. M., Buschek, D., ... & Hussmann, H. (2020). Predicting personality from patterns of behavior collected with smartphones. *Proceedings of the National Academy of Sciences*, 117(30), 17680-17687.

Strogatz. S. (2018) One Giant Step for a Chess-Playing Machine, *The New York Times*, Dec. 26, 2018 https://www.nytimes.com/2018/12/26/science/chess-artificial-intelligence.html

Swanepoel, D. (2020). The possibility of deliberate norm-adherence in AI. *Ethics and Information Technology*. https://doi.org/10.1007/s10676-020-09535-1

Towards a Global Ethics (2018) https://www.weltethos.org/uploaded/documents/wee-2018.pdf

Toyabe, S., Sagawa, T., Ueda, M., Muneyuki, E., & Sano, M. (2010). Experimental demonstration of information-to-energy conversion and validation of the generalized Jarzynski equality. *Nature physics*, 6(12), 988-992.

van de Poel, I.R. (2020). Embedding Values in Artificial Intelligence (AI) Systems. *Minds and Machines*, 30(3), 385-409.

Villaronga, E. F., Kieseberg, P., & Li, T. (2018). Humans forget, machines remember: Artificial intelligence and the right to be forgotten. *Computer Law & Security Review*, 34(2), 304-313.

Youyou, W., Kosinski, M., & Stillwell, D. (2015). Computer-based personality judgments are more accurate than those made by humans. *Proceedings of the National Academy of Sciences*, 112(4), 1036-1040.

Zuse, Konrad (1967), (1969) (1993), Rechnender Raum (Braunschweig: Vieweg). Eng. trans. Calculating Space, *MIT Technical Translation AZT-70–164-GEMIT, Massachusetts Institute of Technology (Project MAC)* (Cambridge, Mass.), 1970.17